\documentclass[10pt,conference]{IEEEtran}

\usepackage{cite}
\ifCLASSINFOpdf
   \usepackage[pdftex]{graphicx}
   \graphicspath{{figs/}}
   \DeclareGraphicsExtensions{.pdf,.jpeg,.png}
\else
   \usepackage[dvips]{graphicx}
   \graphicspath{{../figs/}}
   \DeclareGraphicsExtensions{.eps}
\fi
\usepackage[cmex10]{amsmath}
\usepackage{amsthm}

\usepackage{algorithmic}
\usepackage{array}
\ifCLASSOPTIONcompsoc
  \usepackage[caption=false,font=normalsize,labelfont=sf,textfont=sf]{subfig}
\else
  \usepackage[caption=false,font=footnotesize]{subfig}
\fi
\usepackage{url}
\begin{document}
\title{Automatically Dataset Augmentation Using Virtual Human Simulation}

%------------------------------------------------------------------------- 
% change the % on next lines to produce the final camera-ready version 
\newif\iffinal
%\finalfalse
\finaltrue
%\newcommand{\jemsid}{99999}
%------------------------------------------------------------------------- 


\iffinal

\author{\IEEEauthorblockN{Marcelo C. Ghilardi, Leandro Dihl, Estev\~{a}o Testa, Pedro Braga, Jo\~{a}o P. Pianta, Isabel H. Manssour, Soraia R. Musse}
\IEEEauthorblockA{DaVint - Data Visualization and Interaction Lab, \\ 
 VHLab - Virtual Human Simulation Lab, \\
 School of Technology, PUCRS-Pontifical Catholic University of Rio Grande do Sul,\\
 Porto Alegre, Brazil}
% \and
% \IEEEauthorblockN{Homer Simpson}
% \IEEEauthorblockA{Twentieth Century Fox\\
% Email: homer@thesimpsons.com}
% \and
% \IEEEauthorblockN{James Kirk\\ and Montgomery Scott}
% \IEEEauthorblockA{Starfleet Academy\\
% San Francisco, California 96678--2391\\
% Telephone: (800) 555--1212\\
% Fax: (888) 555--1212}
% \and
% \IEEEauthorblockN{James Kirk\\ and Montgomery Scott}
% \IEEEauthorblockA{Starfleet Academy\\
% San Francisco, California 96678--2391\\
% Telephone: (800) 555--1212\\
% Fax: (888) 555--1212}
% \and
% \IEEEauthorblockN{James Kirk\\ and Montgomery Scott}
% \IEEEauthorblockA{Starfleet Academy\\
% San Francisco, California 96678--2391\\
% Telephone: (800) 555--1212\\
% Fax: (888) 555--1212}
% \and
% \IEEEauthorblockN{James Kirk\\ and Montgomery Scott}
% \IEEEauthorblockA{Starfleet Academy\\
% San Francisco, California 96678--2391\\
% Telephone: (800) 555--1212\\
% Fax: (888) 555--1212}
% \and
% \IEEEauthorblockN{James Kirk\\ and Montgomery Scott}
% \IEEEauthorblockA{Starfleet Academy\\
% San Francisco, California 96678--2391\\
% Telephone: (800) 555--1212\\
% Fax: (888) 555--1212}
}

% SE APROVADO, INCLUIR LISTA DE AUTORES ABAIXO
%\name{Marcelo C. Ghilardi,
%         Leandro Dihl,
%         Estev\~{a}o Testa,
%         Pedro Braga,
%         Jo\~{a}o P. Pianta,
%   Isabel H. Manssour,
%         Soraia R. Musse
% \address{PUCRS-Pontifical Catholic University of Rio Grande do Sul \\
% Graduate Course in Computer Science\\
% Porto Alegre, Brazil\\
% }


%Telephone: (800) 555--1212, Fax: (888) 555--1212}

\else
  \author{Sibgrapi paper ID: 79 }
\fi

\maketitle

\begin{abstract}
Virtual Human Simulation has been widely used for different purposes, such as comfort or accessibility analysis. In this paper, we investigate the possibility of using this type of technique to extend the training datasets of pedestrians to be used with machine learning techniques. Our main goal is to verify if Computer Graphics (CG) images of virtual humans with a simplistic rendering can be efficient in order to augment datasets used for training machine learning methods. In fact, from a machine learning point of view, there is a need to collect and label large datasets for ground truth, which sometimes demands manual annotation. In addition, find out images and videos with real people and also provide ground truth of people detection and counting is not trivial. If CG images, which can have a ground truth automatically generated, can also be used as training in machine learning techniques for pedestrian detection and counting, it can certainly facilitate and optimize the whole process of event detection. In particular, we propose to parametrize virtual humans using a data-driven approach. Results demonstrated that using the extended datasets with CG images outperforms the results when compared to only real images sequences.
\end{abstract}


\IEEEpeerreviewmaketitle

%%%%%%%%%%%%%%%%%%%%%%%%%%%%%%%%%%%%%%%%%%
\section{Introduction}
\label{sec:intro}

In the last years, there is a growing interest in understanding the behavior of pedestrian and crowds in video sequences. It is important in many applications, but certainly one of the most relevant is the safety of pedestrians in complex buildings or in mass events. Many methodologies to detect groups and crowd events have been proposed in the literature and achieved results showing that groups, social behaviors and navigation aspects can be successfully detected in video sequences. For example, counting people in crowds~\cite{Chan2009, cai2014}, abnormal behavior detection~\cite{Ermis2008, Mahadevan2010}, study of social groups in crowds~\cite{Shao2014, Chandran2015}, understanding of group behaviors~\cite{Solmaz2012} and characterization of crowd features~\cite{Zhou2014}. 
Most of these approaches are based on individual pedestrian tracking or optical flow algorithms, and in general consider features like speed, directions, and distance over time. 

On the other hand, many of these applications have been also addressed with another perspective, i.e. by using huge datasets for training and testing machine learning techniques, as described in Section~\ref{sec:related}, in order to reach accurate results. One of the main drawbacks of this area is the needed work to build the datasets and respective ground truth, that mainly for pedestrians and crowds is not a trivial task. Sometimes this ground truth is manually prepared, which is very  time-consuming. Thus, computer graphics simulations started to be used to generate greater labeled datasets to apply as ground truth. LCrowdV~\cite{Cheung2016} is a recent example of computer graphics technology to generate crowds datasets from a set of provided parameters.

In this paper, we intend to investigate the efficiency of CG images generated with our framework to simulate crowds and render Virtual Humans (VH) in a simplistic way. In particular, we are interested about semi-automatically generating CG images based on a labeled dataset, i.e. using data-driven techniques. The idea behind is to use a parameterized crowd simulator, where information from trajectories labeled in the dataset generates parameters for crowds. We used two crowd simulators to simulate virtual humans and generate automatically the ground truth. In addition, images were rendered using the Unity Engine. We also implemented the Multi-column Convolutional Neural Network (MCNN) proposed in~\cite{Zhang2016} to test the CG dataset and compare the efficiency with known and used dataset UCSD~\cite{Chan2008}. The main contribution of this paper is the discussion and investigation of VH simulation used to extend crowds and pedestrian datasets in a data-driven way. Results indicate that this idea is indeed promising since it reduces the work of generating ground truth datasets manually labeled.

This paper is organized as follows: Section~\ref{sec:related} describes the literature review on the topic of crowd counting and VH simulation focused on dataset generation. The proposed model is presented in Section~\ref{sec:model}, while experimental results are discussed in Section~\ref{sec:results}. Finally, Section~\ref{sec:conclusions} draws some conclusions and future work.

%%%%%%%%%%%%%%%%%%%%%%%%%%%%%%%%%%%%%%%%%%
\section{Related Work}
\label{sec:related}

In the last years, several works have been developed for crowd counting~\cite{Chan2009, Felzenszwalb2010, Fiaschi2012, cai2014, Zhang2015, Hu2016} with different purposes, as crowd control, urban planning and video surveillance~\cite{Ma2016}. This problem consists in the definition of the number of people in a crowd~\cite{Sheng2016}, and has been addressed over the years using several approaches~\cite{Viola2005, Rabaud2006, Chan2008, Felzenszwalb2010, Fiaschi2012}, such as Support Vector Machine (SVM) classifier~\cite{Cortes1995} and object detection using a boosted cascade of features~\cite{viola2001}. 

Recently, trying to improve the results accuracy, different methods of Convolutional Neural Networks (CNN) have been widely used~\cite{Zhang2016, Boominathan2016, Walach2016, Gao2016-b, Onoro-Rubio2016, Zhao2016, Sourtzinos2016}. Sourtzinos et al.~\cite{Sourtzinos2016} presented a method for people counting using CNN, and tested with the available Mall crowd counting dataset~\cite{chen2012}. This dataset was annotated manually through the labeling process of head position for each pedestrian in all frames.

Zhang et al.~\cite{Zhang2016}, e.g., proposed a Multi-column Convolutional Neural Network (MCNN) that allows input images of arbitrary resolution. However, besides using existing datasets, they also had to collect and annotate a huge dataset to perform the experiments in order to verify the effectiveness of their method. Another large-scale dataset with annotated pedestrians for crowd counting algorithms was provided by Zhao et al.~\cite{Zhao2016}. Gao et al.~\cite{Gao2016-b} combined the Adaboost algorithm and the CNN for head detection and used a classroom surveillance dataset also manually annotated to evaluate the proposed method. 

Due to the need for this large amount of training data, Boominathan et al.~\cite{Boominathan2016} performed an augmentation of their training dataset cropping patches from the multi-scale pyramidal representation of each training image. Cheung et al.~\cite{Cheung2016}, on the other hand, claim that the task to manually label the datasets is time-consuming and error-prone, besides needing several human operators. Therefore, they proposed a procedural framework called LCrowdV, to generate labeled crowd videos. 

Thus, it is possible to see that although CNN methods present excellent results, there is a need to collect and label large datasets for ground truth, which sometimes demands manual annotation. Because of this, recent research has been addressed to help the problem of generating labeled videos, as the LCrowdV developed by Cheung et al.~\cite{Cheung2016}. Synthetic data has already been used to improve image recognition ~\cite{Thian2003, Zuo2007, Galbally2012}, however, this approach was not yet explored in crowd/pedestrian counting solutions. 

One advantage of crowd simulation applications is the possibility to easily generate a huge dataset together with a ground truth, which fully eliminates the need for an annotated ground truth, and this fact is an important and relevant advantage. This advantage is further enhanced with the possibility to generate automatically labeled crowd videos similar to the real ones, in order to easily extend existent datasets to be used to train machine learning techniques.

%%%%%%%%%%%%%%%%%%%%%%%%%%%%%%%%%%%%%%%%%%
\section{Proposed Model}
\label{sec:model}

Since the focus of this paper is to discuss the automatic process of augmenting labeled datasets, we chose to use one state-of-the-art architecture~\cite{Zhang2016} to conduct our research. We implemented the Multi-column Convolutional Neural Network (MCNN)~\cite{Zhang2016} due to the contribution presented by the authors: their method can manage features at different scales all together in order to accurately estimate crowd counts for different images. However, first of all, we used our simulators in order to simulate virtual humans and generate Virtual Human datasets. Section~\ref{sec:crowds} describes details about this process.

\begin{figure}[h]
\centering
	\includegraphics[width=9cm]{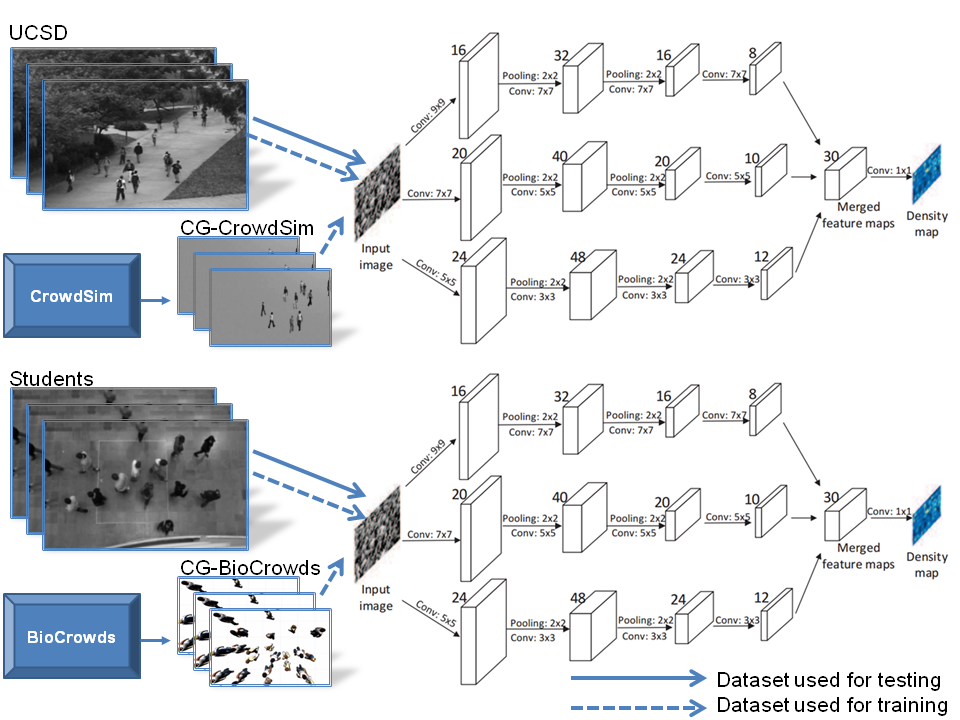}
	\caption{Outline of the proposed methodology. The UCSD and Students datasets are used to train and test, already the CG dataset was used for only training.}
	\label{fig::esquema1}
\end{figure}

The overview of our method is presented in Figure~\ref{fig::esquema1}. It is possible to see the illustration of four used datasets. Firstly, on the top-left appears the UCSD~\cite{Chan2008} images that were used for training and testing. Such dataset contains low dense crowds and ground truth data. The dataset called ''Students'' was filmed in our University and presents from 0 to 30 students in a top-view camera, in an environment of $9$sqm. The goal is to provide a dataset with a different camera perspective as well as different crowd density if compared to UCSD. This dataset was also used to train and test our method. We tracked the people in Students dataset using a method proposed by Bins et al.~\cite{Bins2013}. We visually analyzed all tracking data and manually corrected any possible problem, in order to generate a semi-automatic accurate ground truth for Students too. The other two datasets, ''CG-CrowdSim'' and ''CG-BioCrowds'', were generated through simulation in order to augment the training data used, as explained below.  

In addition, we used the CG images as a dataset for only the training phase in the machine learning method. Trajectories and behaviors have been generated using two simulators (i.e. CrowdSim and BioCrowds) and rendered at Unity. These two simulators are controlled in a different way. While CrowdSim has a graphical interface that can be used to define the simulation someone is interested in, BioCrowds is a parametrized simulator, that can be data-driven. We firstly used CrowdSim because it is more comparable with LCrowdV method~\cite{Cheung2016}, i.e. the crowd designer should manually define initial positions, goals, speeds and etc. So, in CrowdSim case (as developed using LCrowdV) people design a crowd they are interested to simulate. In the case of this work, it is clear that we want to ''imitate'' the dataset, providing the augmentation. However, this imitation process is user-based since the UCSD data is not used as parameters for CrowdSim. More details about CrowdSim is presented in Section~\ref{sec:crowdsim}. In order to test BioCrowds we read the information stored into the dataset Students and provide an automatic parametrization of the simulator, as discussed in Section~\ref{sec:Biocrowds}.

In both cases, we used a very simplistic method at Unity (e.g. virtual humans do not generate shadow on the floor). In order to generate the ground truth for machine learning method (agents positions in image coordinates at a function of time), we used the clear advantages that in both simulators all virtual humans positions are known in world coordinates. So, we assumed a classical pinhole camera model $u = P x$, where $u$ is the pixel in the image (homogeneous coordinates), $P$ is the projection matrix $3 \times 4$ (known in CG world) , e $x$ is the $3D$ position in the world (also inhomogeneous coordinate).
In the next sections, we discuss some details about crowd simulators, and then some information about how density maps were generated to be used in the MCNN.

\subsection{Crowd Simulators}
\label{sec:crowds}

This section details the two used crowd simulators.

\subsubsection{CrowdSim}
\label{sec:crowdsim}

CrowdSim is a rule-based crowd simulation software developed to simulate coherent motion and behaviors of virtual humans in a geometric environment. In particular, CrowdSim has been used to simulate evacuation scenarios~\cite{Cassol:2016}. CrowdSim simulates VH, while keeping behaviors as seek-to-goal and collision avoidance, and also generates outputs to be used in post-processing phases, such as the position of each agent at each time that can be used to visualize the characters in other platforms. In addition, CrowdSim generates statistical data that are used to estimate human comfort and safety in a specific environment, e.g. densities, velocities and etc. 

Two key components are considered in CrowdSim, organized in distinct modules: \textit{Configuration} and \textit{Simulation}, which are respectively responsible for configuring the environment/population/routes information and for the simulation and events.   
For further details, please refer to~\cite{Cassol:2016}. 

\begin{figure}[h]
\centering
	\includegraphics[width=9cm]{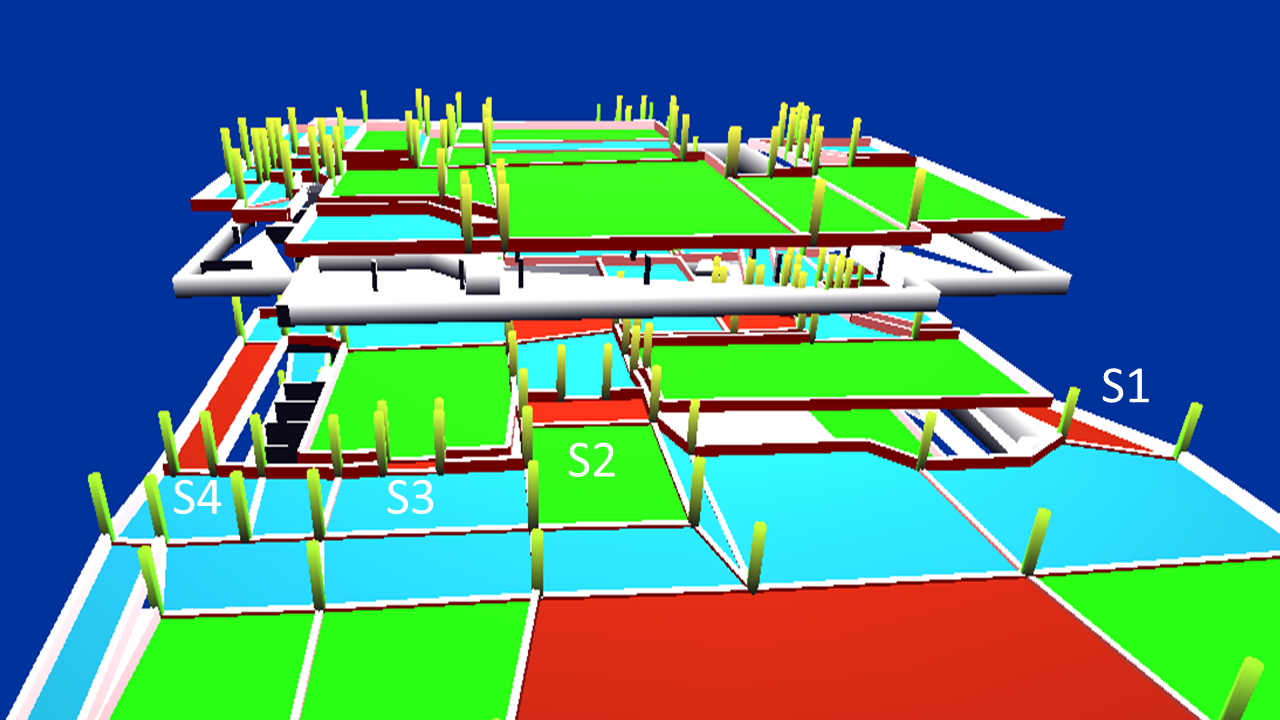}
	\caption{CrowdSim environment example. S1, S2, S3, and S4 represent the exits in a simulated night club.}
	\label{fig::crowdsim}
\end{figure}

Figure~\ref{fig::crowdsim} illustrates CrowdSim environment. We can see the CrowdSim contexts (walkable regions) and connections among them (white edges) that guide agents to the pre-defined exits. S1, S2, S3, and S4 represent the exits in the simulated night club~\cite{Cassol:2016}, that was also simulated in real life. The advantages of CrowdSim, if compared to other crowd simulators in literature, is that it has been evaluated and validated according to Galea~\cite{Galea:1998} and also tested in a real scenario. The navigation graph generated by CrowdSim (edges are routes and contexts are nodes), together with the population distribution in the entry contexts (entry rooms) and the expected distributions at the decision points form the definition of our crowd motion plans.

\subsubsection{BioCrowds}
\label{sec:Biocrowds}

\begin{figure}[t]
\centering
\includegraphics[width=0.8\linewidth]{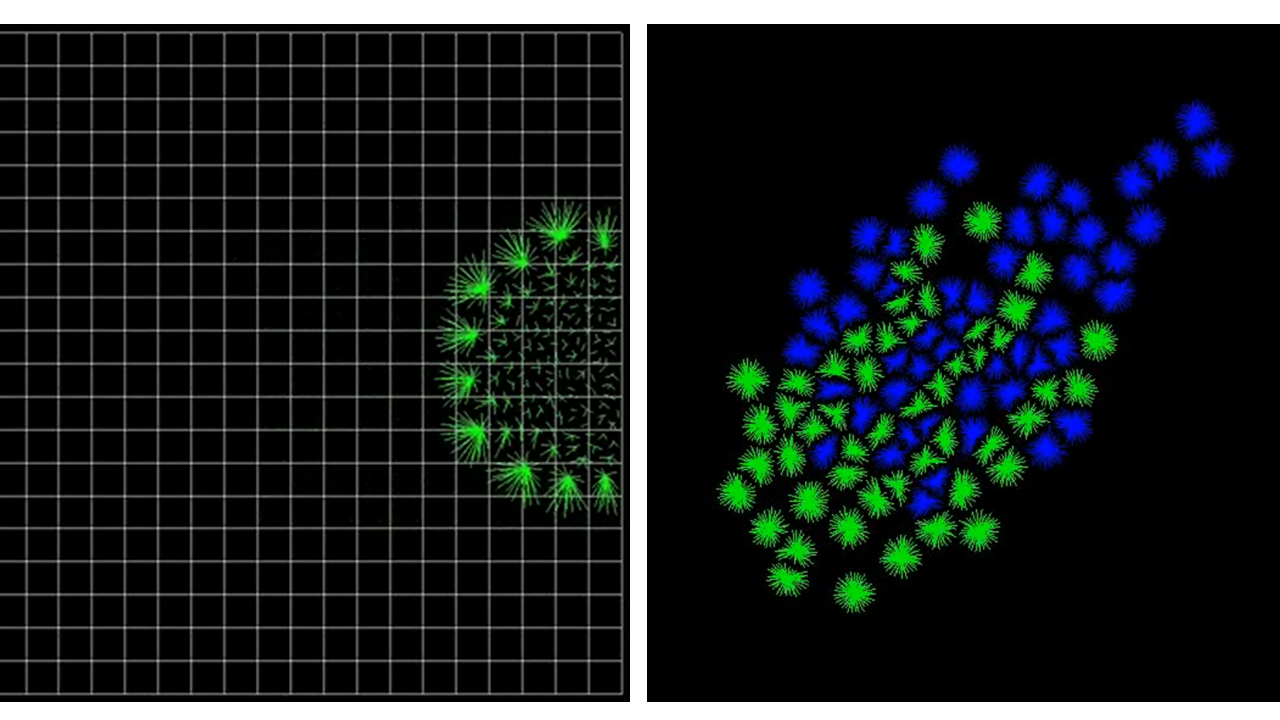}
\caption{BioCrowds: We show emergent phenomena being produced by our simulator in a manner similar to other crowd frameworks; on the left: arc formation and on the right: emergent lanes.} 
\label{fig:emergentBeha}
\end{figure}

One agent in the environment perceives a set of markers (dots) on the ground (described through a space subdivision method) within its observational radius and move forward to its goals taking into account such markers (unoccupied and closest to this agent than any other one). This is the main aspect of BioCrowds simulator~\cite{Bicho12} which supports some of the important emergent behaviors expected in crowd simulation (illustrated in images from Figure~\ref{fig:emergentBeha}), as also emergent in other crowd simulators~\cite{Helbing11,van08a}. As a consequence of BioCrowds main functions, obstacles are very easy to represent as zones without any markers in space discretization method. 
 
In order to provide data-driven control, we read information from the dataset, i.e. the number of individuals $n$ and their positions $x_i^f$ in image coordinates, as a function of time $f$. For the moment we only work with top of view cameras where perspective and coordinates do not need to be transformed. The only performed mapping to generate world coordinates is to find out the correspondent positions in pixels $x_i^f$ to meters $X_i^f$, in order to compute BioCrowds parameters.

We generate following information for each person $i$: speed $s_i^f$ (meters/frame) computed based on $X_i$,  initial  $X_i^{if}$ and final $X_i^{ff}$ positions for each individual, in frames $if$ and $ff$, that represents respectively the first and last frame that individual $i$ appeared in the video sequence. Having this data we are able to parametrize BioCrowds as follows:

\begin{itemize}
\item $n_B = n$, where $n_B$ is the number of agents in BioCrowds;
\item For each agent $j$ in $[0;n_B]$;
\item $X_j^{if} = f(x_i^{if})$, where $i$ is the index of individual in a video sequence and $j$ is the index in BioCrowds and $f(w)$ is a function that maps positions from image coordinates to world coordinates;
\item Similarly, $X_j^{ff} = f(x_i^{ff})$ and $s_j^f$ are computed.
\end{itemize}

Then, BioCrowds is able to simulate the $n_B$ agents having their parameters defined w.r.t input dataset. As output, BioCrowds generates the position of each agent at each frame $X_j^f$. GT and Unity images with virtual humans are the specific output.

It is important to highlight that we chose to simulate people between initial and final frames because we want to be able to simulate the same pattern of crowd existent in the dataset, but allowing to increase or decrease the number of agents, i.e. varying the generated data. For this, we just need to replicate some positions coming from the dataset to serve as input information to agents in BioCrowds. Even if two agents have the same initial and goals positions, they adopt different motion due to the collision avoidance present in BioCrowds method.

\subsection{Generation of Density maps for Computer Graphics Datasets}
\label{sec:density}

As mentioned in~\cite{Zhang2016}, the estimated crowd density, computed from an input image and used in the training step, is very determinant in the CNN performance. In order to provide CG dataset that can be comparable to the UCSD and~\cite{Zhang2016} results, we used the same method, howeve,r adapted to data obtained from virtual human simulation.

Indeed, for CrowdSim dataset we simulated from 0 to 20 agents and their motion aimed to replicate the environment present at UCSD dataset. For BioCrowds dataset we simulated exactly 22 agents and positions from Students were used to people simulation. For both synthetic datasets, the rendering was processed in real time and 30 images were generated per second. Associated to each image, a file was generated having the position $\vec{X}_{i}^f$ of each agent $i$ at each frame $f$ in world coordinates. This set of files was used to transform coordinates from world to image, given the camera position used in CG generation, then generating the position of each agent in image coordinate $\vec{u}_{i}^f$.

\begin{figure}[htb]
\centering
\subfloat[UCSD image.]{\label{fig:UCSD-1}\includegraphics[trim={1.2cm 1.2cm 1.2cm 1.2cm},clip,width=0.3\linewidth]{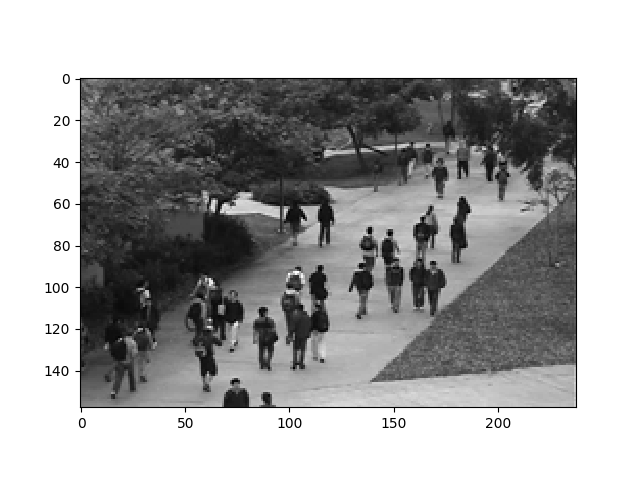}}
\subfloat[UCSD GT.]{\label{fig:UCSD-2}\includegraphics[trim={1.2cm 1.2cm 1.2cm 1.2cm},clip,width=0.3\linewidth]{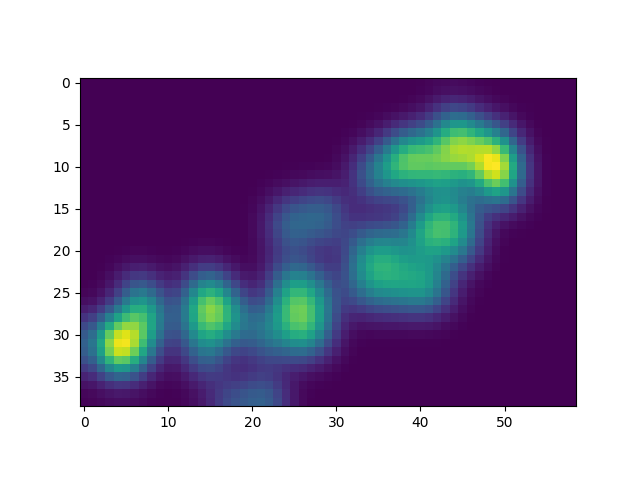}}
\subfloat[UCSD output.]{\label{fig:UCSD-3}\includegraphics[trim={1.2cm 1.2cm 1.2cm 1.2cm},clip,width=0.3\linewidth]{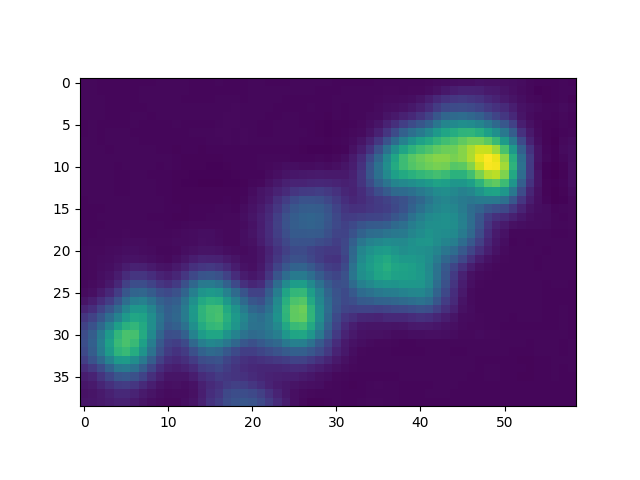}}

\subfloat[CrowdSim image.]{\label{fig:CS-1}\includegraphics[trim={1.2cm 1.2cm 1.2cm 1.2cm},clip,width=0.3\linewidth]{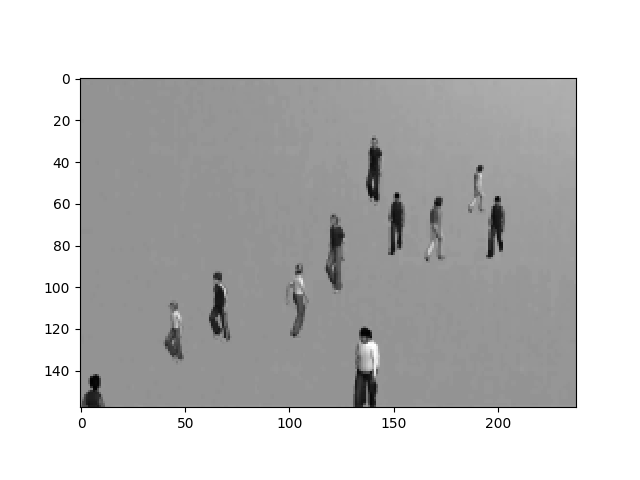}}
\subfloat[CrowdSim GT.]{\label{fig:CS-2}\includegraphics[trim={1.2cm 1.2cm 1.2cm 1.2cm},clip,width=0.3\linewidth]{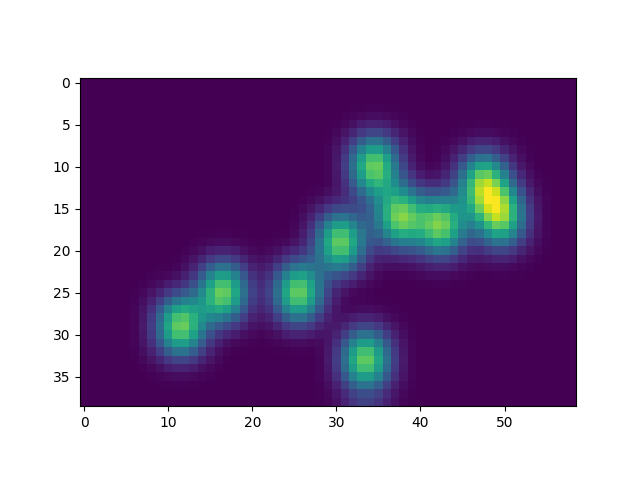}}
\subfloat[CrowdSim output.]{\label{fig:CS-3}\includegraphics[trim={1.2cm 1.2cm 1.2cm 1.2cm},clip,width=0.3\linewidth]{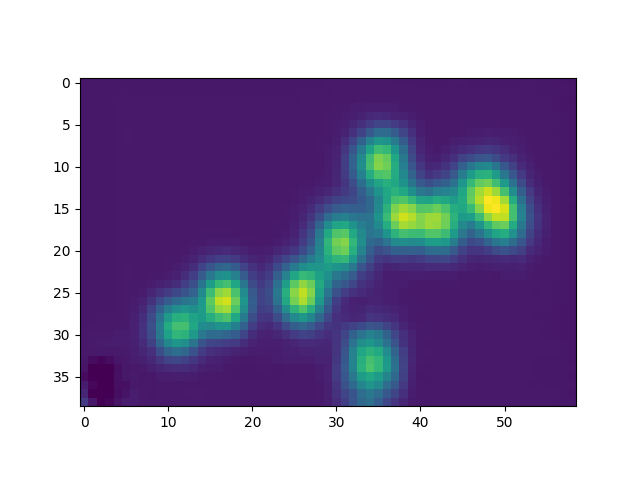}}

\subfloat[Students image.]{\label{fig:Students-1}\includegraphics[trim={1.2cm 1.2cm 1.2cm 1.2cm},clip,width=0.3\linewidth]{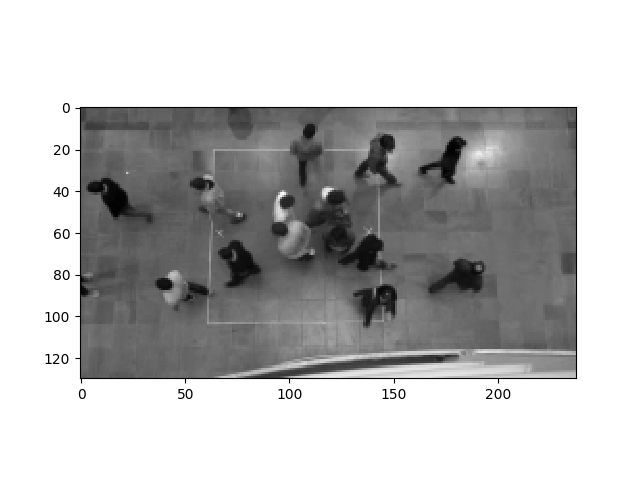}}
\subfloat[Students GT.]{\label{fig:Students-2}\includegraphics[trim={1.2cm 1.2cm 1.2cm 1.2cm},clip,width=0.3\linewidth]{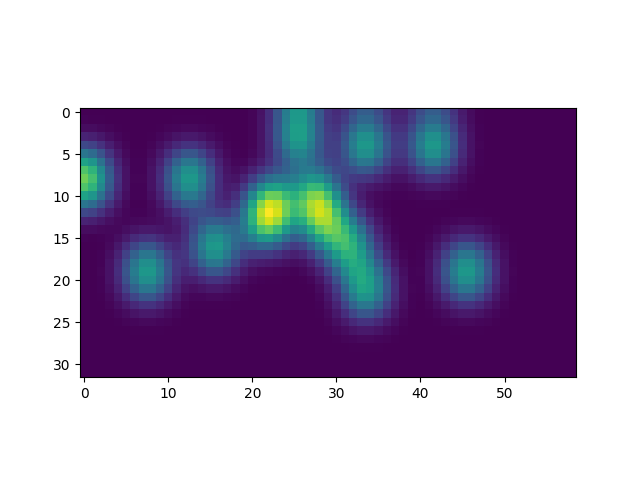}}
\subfloat[Students output.]{\label{fig:Students-3}\includegraphics[trim={1.2cm 1.2cm 1.2cm 1.2cm},clip,width=0.3\linewidth]{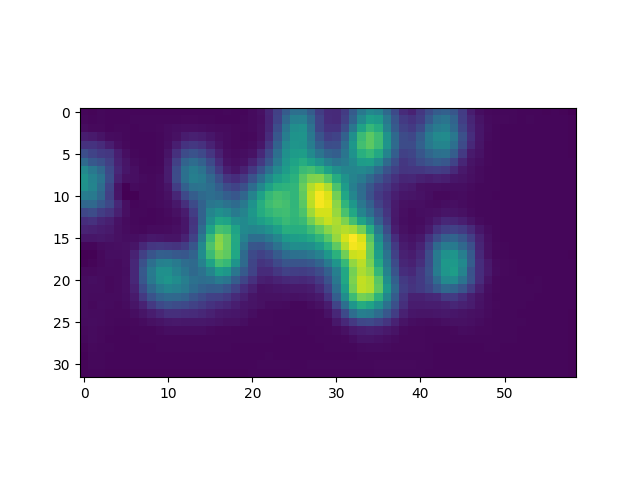}}

\subfloat[BioCrowds image.]{\label{fig:Students-1}\includegraphics[trim={1.2cm 1.2cm 1.2cm 1.2cm},clip,width=0.3\linewidth]{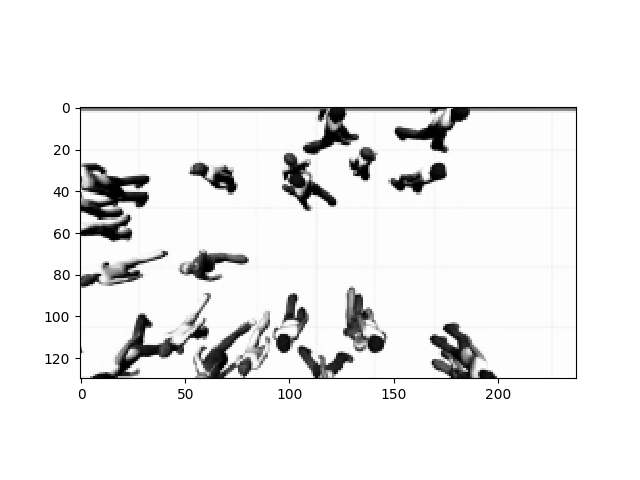}}
\subfloat[BioCrowds GT.]{\label{fig:Students-2}\includegraphics[trim={1.2cm 1.2cm 1.2cm 1.2cm},clip,width=0.3\linewidth]{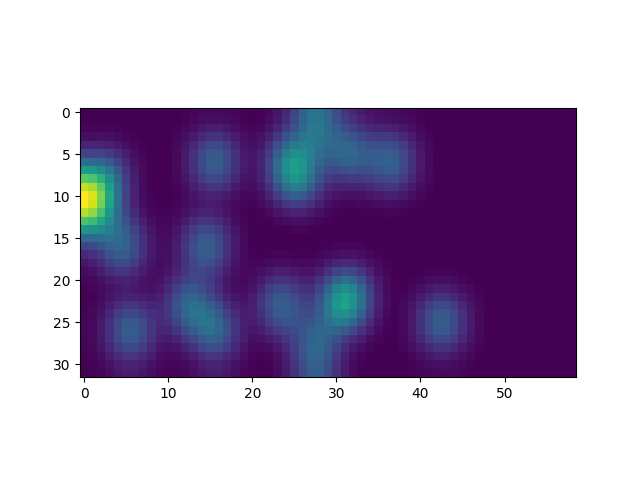}}
\subfloat[BioCrowds output.]{\label{fig:Students-3}\includegraphics[trim={1.2cm 1.2cm 1.2cm 1.2cm},clip,width=0.3\linewidth]{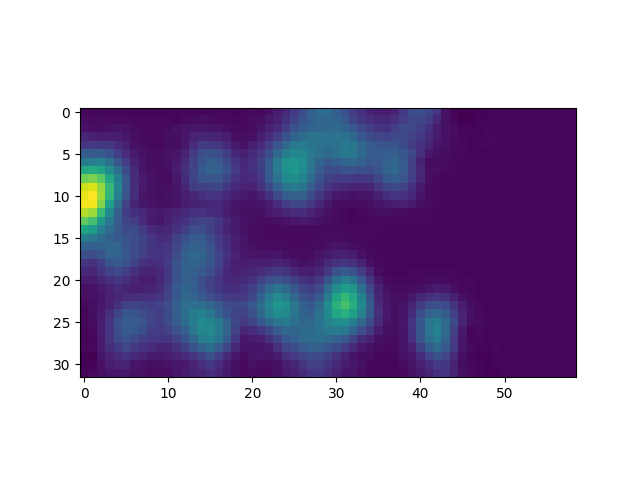}}

\caption{Some images from the UCSD, CrowdSim, Students and BioCrowds datasets. (a), (b) and (c) illustrate the original image, the processed ground truth, and MCNN output, respectively. (d), (e) and (f) are correspondent images for a dataset using computer graphics (CrowdSim). (g), (h) and (i) are correspondent images for Students dataset. (j), (k) and (l) are correspondent images for a dataset using computer graphics (BioCrowds). } 

\label{fig:density map}
\end{figure}

In order to generate the density maps for CG datasets, we use distances among agents in the frame. We denote the distances from agent $i$
to its $k$ nearest neighbors (in image coordinates) as ${d_i^1, d_i^2, ..., d_i^k}$ and the average distance is $\bar{d_i}$. Therefore, to estimate the crowd density around the pixel $\vec{u}_i$, we perform a convolution $\delta{}(\vec{u}-\vec{u}_i)$ with a Gaussian kernel with variance $\gamma_i$ proportional to $\bar{d_i}$. For more details please refer to~\cite{Zhang2016}. Figure~\ref{fig:density map} illustrates images from the four datasets (on the left), the result of density maps generated for GT (middle) and the result of MCNN (right). 

%%%%%%%%%%%%%%%%%%%%%%%%%%%%%%%%%%%%%%%%%%
\section{Experimental Results}
\label{sec:results}

In order to evaluate the performance of CG images in the training phase, we tested two sets of Real Dataset + CG images. Next section (\ref{sec:ucsd+crowdsim}) aims to discuss the results of UCSD augmented with crowd simulation manually defined. Section~\ref{sec:students+biocrowds} has the goal to show the results when a dataset Students was augmented. 

The total number of images are $4630$, being $2000$ from UCSD, $1545$ from CrowdSim (using UCSD as basis), $350$ from Students (filmed in our University) and $735$ from BioCrowds datasets (using Students as basis). The experimental results aim to show that even these simplistic rendering applied in CrowdSim and BioCrowds datasets can improve the performance of machine learning method. As commonly used, we adopted the MAE (Mean Absolute Error) and MSE (Mean Squared Error) metrics.\footnote{Differently from ~\cite{Zhang2016}, we used MSE and not RMSE.}

First, we evaluated individually the four datasets used in this work for training and testing (see  Table~\ref{tab:datasets}). It is easy to see that the MCNN performance works better in CrowdSim and BioCrowds datasets, when compared to others. One difference among the datasets is that the CG images are more homogeneous, given the synthetic background (see Figure~\ref{fig:density map} for illustration about the datasets).

\begin{table}[htb]
\centering
\begin{tabular}{|p{0.3\linewidth}|l|l|l|l|}
\hline
Training    & Testing    & MAE     & MSE \\  \hline
% & & & \\
UCSD(40\%)  & UCSD(60\%)  & 1.3745	 & 9.4633	\\ \hline
CrowdSim(40\%)  & CrowdSim(60\%)  & 0.7898	 & 3,4966	\\ \hline
Students(40\%)  & Students(60\%)  & 1.1087	 & 5.5069	\\ \hline
% Marcelo - 12/06/2017 - Adicionei o valor
BioCrowds(40\%)  & BioCrowds(60\%)  & 1.0981 & 5.5776 \\ \hline
\end{tabular}
\quad
\caption{MCNN applied to the four analyzed datasets.}
\label{tab:datasets}
\end{table}

Next sections describe the performed analysis in the augmented datasets.

\subsection{UCSD and CrowdSim}
\label{sec:ucsd+crowdsim}

\begin{figure}[htb]
  \centering
  \includegraphics[width=0.5\textwidth]{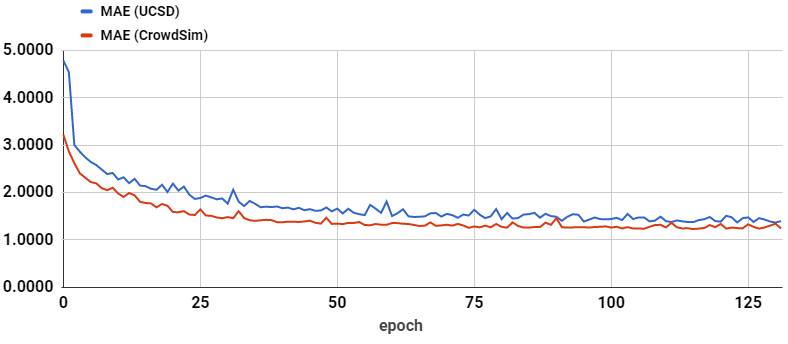}
  \caption{MCNN performance when training with UCSD and when we extended the dataset with CG images generated using CrowdSim.}
  \label{fig:evolution}
\end{figure}

We compared the evolution of MCNN performance in two different situations: \textit{i)} when training only with UCSD (40\%) and \textit{ii)} extending the training dataset with CG images from CrowdSim (from 0\% to 100\% of total 1545 images). For UCSD we used the same
setting than~\cite{Zhang2016}, i.e. we use frames
from 601 to 1400 as training data, and the remaining 1200 frames as test data. For CrowdSim dataset, we selected randomly the images until complete the required percentage  of images in the dataset (from 0\% to 100\% of total 1545 images), to be used as training information. The tested images were always the same set of UCSD (60\%) for both evaluations. Figure~\ref{fig:evolution} shows the evolution in terms of computed epochs. It is easy to see that the extended dataset using CG images improved the MCNN performance.

%In other experiments, we computed MAE and MSE when using three configurations for training phase: \textit{i)} only with UCSD (40\%), \textit{ii)} extended training dataset with CG images (from 20\% to 100\%) and \textit{iii)} extended training dataset with Students dataset (from 20\% to 100\%). 

\begin{figure}[htb]
  \centering
  \includegraphics[width=0.45\textwidth]{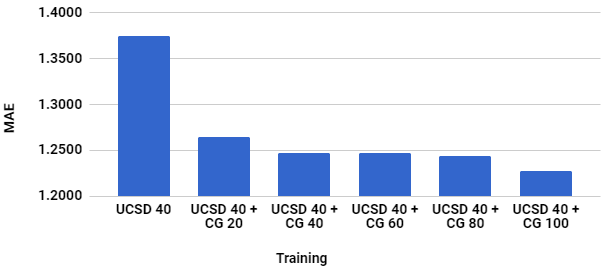}
  \caption{Comparing the MAE metric when training dataset UCSD was extended with CG images generated using CrowdSim.}
  \label{fig:MAE}
\end{figure}

\begin{figure}[htb]
  \centering
  \includegraphics[width=0.45\textwidth]{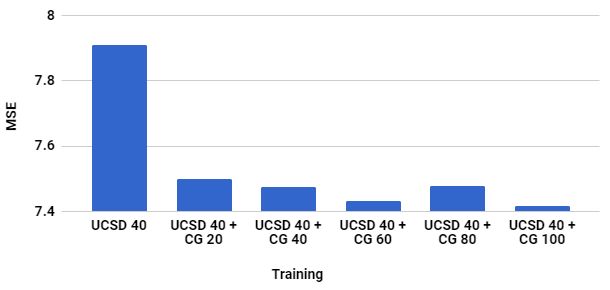}
  \caption{Comparing the MSE metric when training dataset UCSD was extended with CG images generated using CrowdSim.}
  \label{fig:MSE}
\end{figure}

Figures~\ref{fig:MAE} and~\ref{fig:MSE} illustrate the results. It is easy to see that the augmentation in the training dataset with CG images provided significantly better performance than without any extension. 
 %Moreover, it is interesting to see that the performance of the two extended datasets (Students and CrowdSim) were very similar, indicating that the training with CG images can be compatible with real images extension.

In addition, we computed the numerical improvement between the performance with and without augmentation in the UCSD dataset (see Table~\ref{tab:UCSD+CG}). Considering the total images of CrowdSim the improvement was approximately 11\% in comparison to the original non-augmented dataset UCSD.

\begin{table}[htb]
\centering
\begin{tabular}{|l|l|l|}
\hline
\textbf{Training Dataset}   & \textbf{MAE Values} & \textbf{\% improvement} \\  \hline
UCSD 40\% 	& 1.3745	& - \\ \hline
UCSD 40\% + CG 20\% &	1.2649	& 7.9745 \\ \hline
UCSD 40\% + CG 40\% &		1.2468	&	9.2914 \\ \hline
UCSD 40\% + CG 60\% &		1.2467	&	9.2987 \\ \hline
UCSD 40\% + CG 80\% &		1.2438	&	9.5097 \\ \hline
UCSD 40\% + CG 100\% &		1.2279	&	10.6665 \\ \hline
\end{tabular}
\quad
\caption{Comparison of performance when UCSD was extended with CG images.}
\label{tab:UCSD+CG}
\end{table}

% \begin{table}[htb]
% \centering
% \begin{tabular}{|p{0.6\linewidth}|l|l|}
% \hline
% \textbf{Mean MAE differences}   & \textbf{Values} \\  \hline
% UCSD and UCSD+Student  & 0.1118	\\ \hline
% UCSD and UCSD+CrowdSim  & 0.1257		\\ \hline \hline
% \textbf{Mean MSE differences }   & \textbf{Values} \\  \hline
% UCSD and UCSD+Student  & 1.1132	\\ \hline
% UCSD and UCSD+CrowdSim  & 	1.1100	\\ \hline 
% \end{tabular}
% \quad
% \caption{Comparison of performance when UCSD was extended with other datasets (Students and CrowdSim).}
% \label{tab:datasets}
% \end{table}

% When extended the testing dataset (60\% UCSD + 60\% Students), the results demonstrated that training the network with UCSD(40\%) and CG(100\%) obtained better results (MAE $1.3109$ and MSE $7.5102$) than training with UCSD(40\%) more Students(40\%) (MAE $1.3513$ and MSE $8.4196$).

As expected, training the MCNN using only CrowdSim dataset was not efficient for testing with any real dataset, indicating that it is necessary to add some real images to the training to obtain better results. The tests used 100\% of samples from all datasets (Table~\ref{tbl:CG-Only}).

\begin{table}[htb]
\centering
\begin{tabular}{|l|r|r|}
\hline
\textbf{Test dataset} & \textbf{MAE} & \textbf{MSE} \\ \hline
UCSD           & 4.3384       & 43.2086      \\ \hline
Students       & 1.4851       & 8.3711       \\ \hline
% Marcelo - 12/06 - Retirei
%UCSD+Students  & 3.0220       & 33.5501      \\ \hline
\end{tabular}
\caption{Training the MCNN using only CrowdSim dataset and testing with real datasets.}
\label{tbl:CG-Only}
\end{table}

\subsection{Students and BioCrowds}
\label{sec:students+biocrowds}

\begin{figure}[htb]
  \centering
  \includegraphics[width=0.5\textwidth]{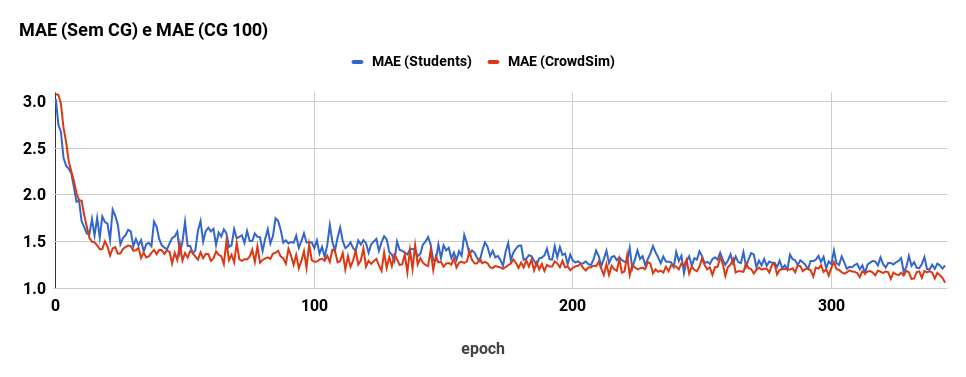}
  \caption{MCNN performance when training with Students and when we extended the dataset with CG images generated using BioCrowds.}
  \label{fig:evolution2}
\end{figure}

As in the last section, we compared the evolution of MCNN performance in two different situations: \textit{i)} when training only with Students (40\%) and \textit{ii)} extending the training dataset with CG images from BioCrowds (from 0\% to 100\% of total 735 images). For both cases, we randomly selected the used images. Figure~\ref{fig:evolution2} shows the evolution in terms of computed epochs. It is easy to see again that the extended dataset using CG images improved the MCNN performance.

%SORAIA: Substituir....
%Marcelo - 12/06/2017 - Substituido

\begin{figure}[htb]
  \centering
  \includegraphics[width=0.45\textwidth]{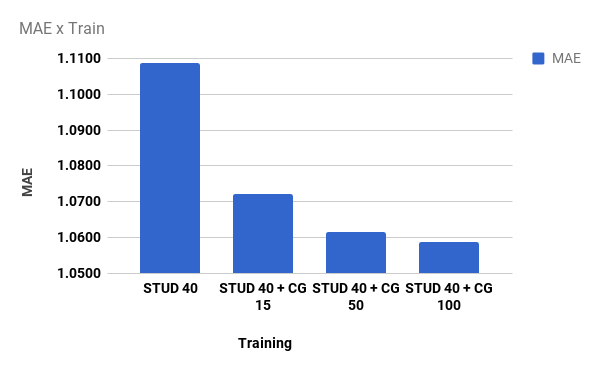}
  \caption{Comparing the MAE metric when training dataset Students was extended with CG images generated using BioCrowds.}
  \label{fig:MAE2}
\end{figure}

\begin{figure}[htb]
  \centering
  \includegraphics[width=0.45\textwidth]{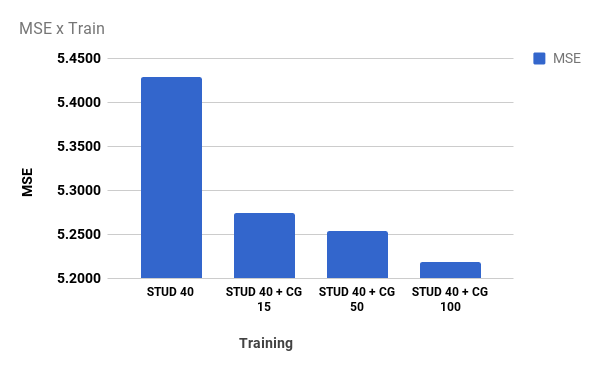}
  \caption{Comparing the MSE metric when training dataset Students was extended with CG images generated using BioCrowds.}
  \label{fig:MSE2}
\end{figure}

Figures~\ref{fig:MAE2} and~\ref{fig:MSE2} illustrate the results. It is easy to see that the augmentation in the training dataset with CG images provided significantly better performance than without any extension. 

%SORAIA: substituir
%Marcelo - 12/06/2017 - Substituido

%SORAIA: substituir
%Marcelo - 12/06/2017 - Substituido

We also computed the numerical improvement between the performance with and without augmentation in the Students dataset (see Table~\ref{tab:datasets2}).

\begin{table}[htb]
\centering
\begin{tabular}{|l|l|l|}
\hline
\textbf{Training Dataset}   & \textbf{MAE Values} & \textbf{\% improvement} \\  \hline
Students 40\% 	& 1.1087	& - \\ \hline
Students  40\% + CG 20\% &	1.0722	&  3.2921 \\ \hline
Students  40\% + CG 40\% &		1.0615	&	4.2572 \\ \hline
Students  40\% + CG 60\% &		1.0588	&	4.5007 \\ \hline
Students  40\% + CG 80\% &		1.0613	&	6.3768 \\ \hline
Students  40\% + CG 100\% &		1.0342	&	7.6305 \\ \hline
\end{tabular}
\quad
\caption{Comparison of performance when Students was extended with CG images. The last column presents the improvement in comparison to non-augmented data.}
\label{tab:datasets2}
\end{table}

Next section presents some final considerations about this work.

%%%%%%%%%%%%%%%%%%%%%%%%%%%%%%%%%%%%%%%%%%
\section{Final Considerations}
\label{sec:conclusions}

In this paper, we described the results of using images of Virtual Human Simulation to extend datasets of pedestrians in order to train machine learning techniques for VH counting and detection.
In order to evaluate our proposal, we implemented the Multi-column Convolutional Neural Network (MCNN) presented by Zhang~\cite{Zhang2016}.
We used four datasets: UCSD~\cite{Chan2008}, also used in paper~\cite{Zhang2016}; a new one called Student, recorded in our University; a synthetic one created with virtual human simulation similar to UCSD dataset and another synthetic dataset based on Students generated in an automatic way using BioCrowds. 

We trained the MCNN with different samples of the datasets: UCSD only, UCSD+CG, Students only and Students+CG.
The results after training with our extended datasets outperform the results of training using just the original dataset, i.e., there was an increase in network performance using the CG extended datasets. In particular, augmenting UCSD with CG images we obtained approximately 10\% of improvement in MAE values, while the improvement in Students was approximately 7\%. These results were coherent with LCrowdV information when the authors said that their improvement is around 7\%. Of course, this number depends on the characteristics of the augmented dataset. 
The performance improved for both tested datasets demonstrates the good generalization of the proposed investigation. Moreover, the possibility of automatically generating a ground truth for labeling datasets facilitates and optimizes the process of pedestrian detection and counting, decreasing the arduous task of manually labeling the videos. 

We also tested two crowd simulators which main difference was the way to control the animations. While in CrowdSim we manually designed experiments imitating UCSD dataset, BioCrowds was automatically parametrized based on datasets. Although tests are necessary, the two simulators do not present differences, in the learning process, since rendering and humans visualization are in the same platform. The only difference is the required work associated with the task to animate crowds, much more easier if using BioCrowds.

For future work, we intend to create, evaluate and let available new datasets of CG images simulating several sizes and densities of crowds. We also want to provide extended datasets to other known datasets, such as the Mall crowd counting dataset~\cite{chen2012}.


% \section*{Acknowledgment}
% Thanks to Brazilian agencies: CAPES, CNPQ and FAPERGS. 
% SE APROVADO INCLUIR AGRADECIMENTOS DO BEPiD!!!
%This work was partially supported by grants from BEPiD/PUCRS (Brazilian Education Program for iOS Development).



\bibliographystyle{IEEEtran}
\bibliography{example}
%\begin{thebibliography}{1}

%\end{thebibliography}

\end{document}